\documentclass{llncs}
\usepackage{amsfonts}
\usepackage{amssymb}
\usepackage{tikz}
\usepackage{graphicx}
\usepackage{qtree}
\usepackage{gb4e}

\begin{document}

\title{Lexpresso: a Controlled Natural Language}
\author{Adam Saulwick}
\institute{Defence Science \& Technology Organisation, Australia}
\date{February, 2014}
\maketitle

\begin{abstract}
This paper presents an overview of `Lexpresso', a Controlled Natural Language developed at the Defence Science \& Technology Organisation as a bidirectional natural language interface to a high-level information fusion system. The paper describes Lexpresso's main features including lexical coverage, expressiveness and range of linguistic syntactic and semantic structures. It also touches on its tight integration with a formal semantic formalism and tentatively classifies it against the PENS system.

\keywords{Controlled Natural Language, Formal Semantics, Linguistic structures, Human--Computer Interaction, High-level Information Fusion}
\end{abstract}

\section{Introduction}
\label{intro}
`Lexpresso' is a Controlled Natural Language (CNL) developed at the Defence Science \& Technology Organisation as a bidirectional natural language interface to a high-level, agent-based, information fusion system called Consensus. This paper is the first published description of Lexpresso's broad features, including lexical coverage, and range of syntactic and semantic structures. It also describes the tight integration with DSTO's bespoke formal semantic formalism, Mephisto, initially conceived by Lambert \& Nowak \cite{Lambert2008}. Lexpresso was first developed in 2008 and is under active development.\footnote{Some previous work in natural language interfaces at DSTO focussed on automated speech-to-text recognition linked to template rules. From June 2007 to July 2008 a DSTO initiated collaborative research program into situation awareness was conducted between DSTO and NICTA, see~\cite{Baader2009}. Among other things, this research involved the development of a CNL which was based on, or inspired by, PENG \cite{Schwitter2004,Schwitter2007}. Subsequent to these activities Lexpresso was built from scratch by DSTO.}

The Consensus system performs high-level information fusion of heterogeneous data for Situation Awareness. In our current demonstration system, synthesised input data types include maritime and aviation tracks\footnote{Synthesised tracks are currently processed at circa 100 per second and contain fields for source, temporal offset, track ID, time, coordinates, direction, speed, class, type, allegiance and nationality.}, natural English texts, emails and spoken English statements. In general terms the Consensus system is designed to demonstrate a working solution to problems of high-level information fusion by the `semiautomation of [some of] the functionalities of sensation, perception, cognition, comprehension, and projection that [are] otherwise performed by people for situation awareness'~\cite{Lambert2012a}. Among other functions Consensus does this by automatically transforming diverse information sources into a canonical semantic machine-readable form called Mephisto which facilitates computational reasoning. A problem however is that Mephisto is only interpretable by machines or by a few human experts (and then slowly) and so, human interaction with Consensus would be very difficult, if not impossible, even for specialists, without a natural interface. Lexpresso is that interface. It bridges the natural-language/formal-language gulf and thus it permits relatively natural interaction with a formal semantic reasoning system via spoken and written controlled natural English.\footnote{Consensus also utilises other interfaces such as a 3-dimensional geospatial display and a virtual adviser avatar. These are not discussed here.}

Lexpresso's bidirectionality means that it has both input and generation capabilities. Further, because Lexpresso is tightly coupled with Consensus's formal semantic knowledge representation and reasoning system whose primary function is automated inferencing over real-time track data and texts for enhanced situation awareness, it provides human users with the enhanced ability to query the nature of current and historical real-world and potentially far-flung events. Answers are given in the form of situation reports. These reports may concern the transit or spatiotemporal interaction of observed maritime, land \&/or air-based platforms and even the social relationships between people inferred from certain text descriptions. 

While these capability descriptions are accurate, they are not intended to obfuscate Lexpresso's limitations. Its breadth and degree of coupling with Mephisto are the subject of ongoing research and development. Consensus is a prototype system and, subject to space constraints, some limitations will be mentioned in Section~\ref{sem-struc}.

The remainder of this paper is structured as follows. Section~\ref{sys-arch} describes the system architecture and the main CNL modules. Sections~\ref{synt-struc} and \ref{sem-struc} exemplify the main syntactic and semantic structures respectively. Section~\ref{classif-lexpresso} proposes a classification of Lexpresso based on the PENS system. Section~\ref{concl} concludes with a summary. 

\section{System architecture \& module functions}
\label{sys-arch}
Echoing the traditional transformational grammar distinction~\cite{Chomsky1957,Chomsky1965}, language processing in Lexpresso is conceived on a spatial metaphor of depth in which `surface CNL' refers to observed spoken or written forms of language and `deep CNL' refers to an underlying abstraction with certain linguistic features. Input processing takes surface language and transforms it into deep linguistic structures. After further processing to remove ambiguities, these are converted into our universal semantic representations called Mephisto structures. Reasoning and inferencing is primarily performed on Mephisto structures. Output processing takes Mephisto structures and uses the same core syntactic parser to validate and generate surface CNL for consumption by users.

\begin{figure}[ht]
	\centering
		\includegraphics[scale=0.59]{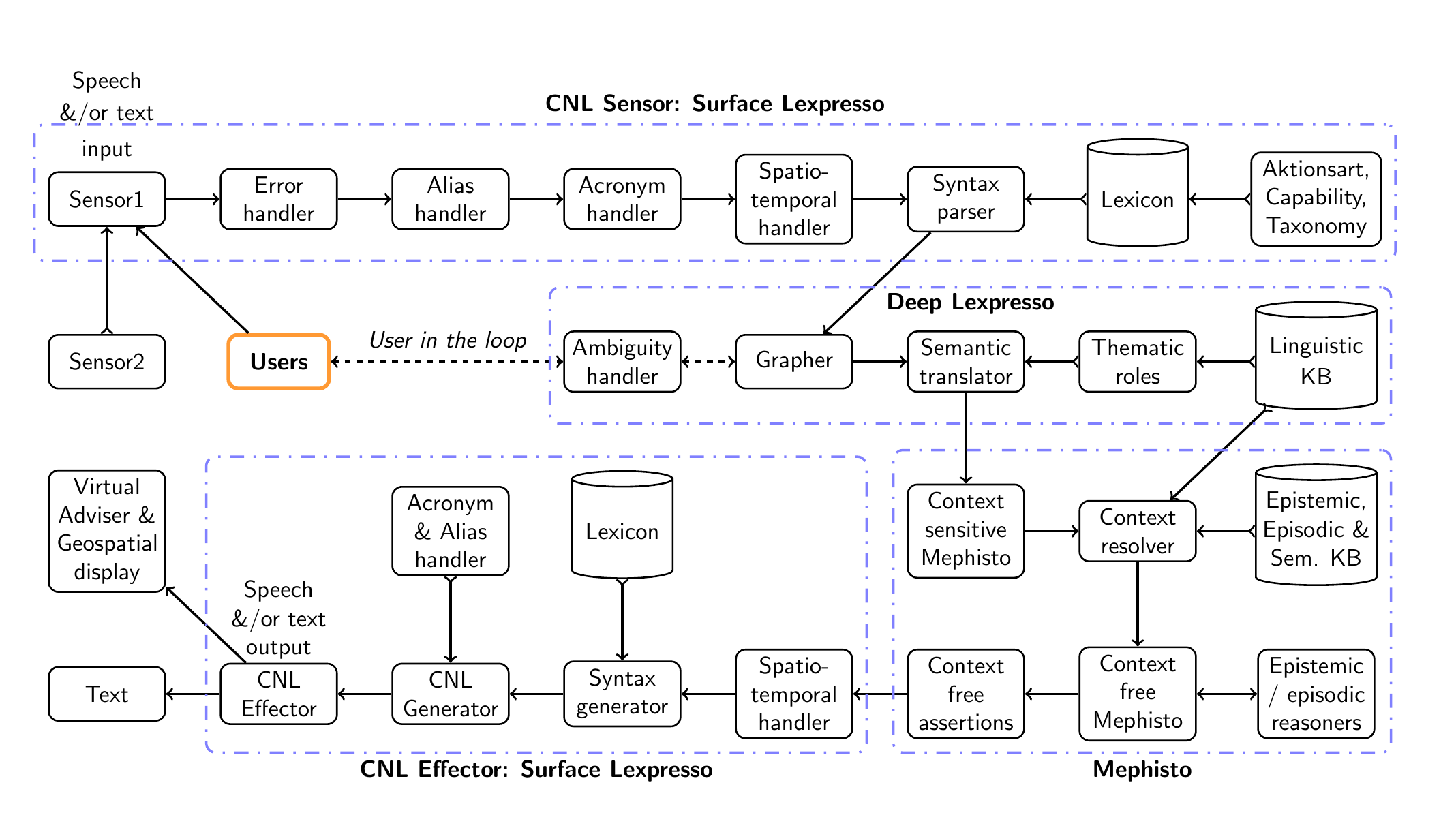}
	\caption{Lexpresso system architecture}
	\label{fig:Lexpresso-system-diagram-v3}
\end{figure}

Lexpresso is designed as a modular system to facilitate integration of new features as required. Depending on how one counts them, it consists of around 17 modules, see Figure \ref{fig:Lexpresso-system-diagram-v3}. Due to space constraints not all components are described.

\vspace{-1cm}
\vspace{-0.8cm}

\begin{figure}[ht] 
	\centering
		\includegraphics[scale=0.4]{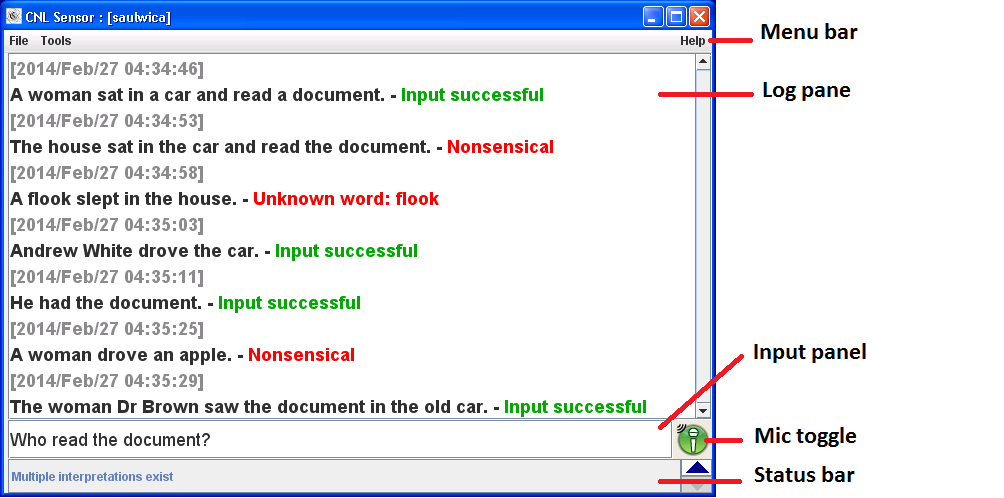}
	\caption{CNL Sensor: showing sample text (with timestamps, proper names, person title \& anaphoric resolution), input panel (with possible query), colour-coded feedback messages in log pane, microphone toggle button (on) for speech input \& status message}
	\label{fig:Error-handler6}
\end{figure}

\paragraph{CNL Sensor} This is the users' primary input screen. It contains a text panel for typing controlled natural English. Spoken English also appears in this panel via the Automated Speech Recogniser. Sentences can be automatically or manually inserted. Manual insertion is an interactive process and thus permits error correction prior to further processing. Once inserted the input appears in the CNL Sensor log window accompanied by the name of the `teller' and a status message. Each line is also timestamped, displayed directly above it, see Figure~\ref{fig:Error-handler6}.\footnote{The `sensor' \& `effector' terminology is adopted from the Attitude Too cognitive model~\cite{Lambert2012}.}

\paragraph{Error handler} During manual insertion, a pre-parser checker notifies the user of unknown or undefined words or out-of-grammar expressions. This provides dynamic feedback on lexical coverage and grammaticality of surface input to inform the user of input status.

\paragraph{Alias handler} This module converts particular multi-word expressions into atomic terms for manipulation at the deep linguistic level, e.g. \texttt{`Becker',`Bender', air, force, base} becomes \texttt{becker\_bender\_AFB}. It is also used to handle contractions, e.g. `can't' becomes `cannot'. It is also used for mapping fixed idiomatic forms to a single lexical correspondence. Although the functionality of the aliasing module is currently used for simple surface level structures, it is also capable of handling metonymy.

\paragraph{Acronym handler} This module expands acronyms and titles into multi-word expressions. It also constrains their syntactic position (e.g. pre-/post-nominal) based on their part-of-speech.

\vspace{-0.8cm}

\paragraph{Spatiotemporal handler} This module converts date and time information into universal standard timestamps, accepting a broad range of natural language expressions such as time-formatted numerals (e.g. 13:59:59 or Zulu time, and a variety of time points in natural language, e.g. 1 PM, one o'clock). Each time is calculated against Coordinated Universal Time (UTC). Our system is designed to allow the sensor, effector and cognition modules to be in different spatiotemporal locations, hence the generation of temporal phrases is calculated on an off-set to UTC.

Temporal intervals are also handled. Surface level forms include the template `from \textsc{time} to \textsc{time}', and a range of temporal words, including `today', days of the week, months of the year, decades, and centuries, etc. Where required by natural English, these  phrases can be further modified with prepositions, such as `from 12:00 to 13:00', `in January', `for a week', and with temporal grounding to specific times, such as `last week', `yesterday', `in one month'. Inferencing with these temporal expressions as reference points is done in the cognition using Allen interval algebra \cite{Allen1983}, not discussed further here. 

Lexpresso pays careful attention to the subtleties of the English tense system. For instance the relationship between the time of user interaction and the tense of the assertion or query is captured and stored as temporal information relevant to each entity. To achieve this, each input is internally labelled with utterance type and time. The latter becomes the reference time for every CNL interaction. For instance, in (\ref{sensor-interaction}),\footnote{Our notation uses a ternary @-predicate---adopted from the Mephisto conceptualisation with a perdurantist philosophy which `holds that an identity is formed from different things at different times, that an identity is a process, an assembly of different temporal parts'---representing @(label, time, space). (Space constraints prohibit discussion of this, but see~\cite{Lambert2008}.) The first argument is the entity's identifier (here a Skolem constant), the second is a temporal point, interval or even a sum of temporal intervals, and the third is the space it occupies. See \cite{Saulwick2013} for further details.} via a \texttt{tells} predicate, the system registers that it has perceived a CNL Sensor interaction at a certain time and from a certain teller; here the author.

\begin{exe}
\ex\label{sensor-interaction}
\begin{small}\begin{verbatim}
	Mon Jun 02 10:33:48 CST 2014 [SENSOR : INTERACTION] 
	perceive(cnl_sensor,tells(teller(@(skc1,invl
	(timestamp(2014,6,2,1,3,48),timestamp(2014,6,2,1,3,48)),s_5),
	Adam_Saulwick),...)),\end{verbatim}\end{small}
\end{exe}

Example (\ref{stood-in-house}) demonstrates how the system stores spatiotemporal information about entities.\footnote{Abbreviations to the examples indicate representations at Input (I), Cognition (C), and Output (O) levels. Skolem constant numbers have been simplified for expository purposes but numbers are automatically assigned at input. Further, information not germane has been omitted and replaced by ellipses. Space limits a full description of the contents of C here.} Given the simple past tense of the surface input sentence in (\ref{stood-in-house-i}), the time of `standing' is encoded at the deep Mephisto cognitive level (\ref{stood-in-house-c}) to have taken place \emph{before} the time of the assertion registered in (\ref{sensor-interaction}). 

\begin{exe}
\ex\label{stood-in-house}
	\begin{xlist}
	\ex[\label{stood-in-house-i}I:]{The woman stood in the house.}
	\ex[\label{stood-in-house-c}C:]{\small\texttt{animate(@(skc2,t\_4,s\_2)),female(@(skc2,t\_4,s\_2)),}\\
	\small\texttt{before(t\_4,invl(timestamp(2014,6,2,1,3,48),}\\
	\small\texttt{timestamp(2014,6,2,1,3,48))),}\\
	\small\texttt{location\_in([stands(@(skc2,t\_4,s\_2))],@(skc3,t\_4,s\_3)),}\\
	\small\texttt{woman(@(skc2,t\_4,s\_2),[animate,definite,singular,...]),}\\ 
	\small\texttt{house(@(skc3,t\_4,s\_3),[definite,singular,prep(in)]),}\\
	\small\texttt{stands[@(skc2,t\_4,s\_2)],[past,...])).} 
  }
  \ex[\label{stood-in-house-o}O:]{The woman stood in the house before Monday the 2nd of June 2014 at 10:33:48 AM.}
	\end{xlist}
\end{exe}

For explicitness by default the time before which the event is asserted to have occurred is rendered visible to the user in the CNL Effector window, as in (\ref{stood-in-house-o}).\footnote{A number of the other underlying linguistic features of the surface input are also sent to the cognition---namely animacy, gender, part-of-speech, number, and definiteness---as exemplified for the intransitive simple past phrase in (\ref{stood-in-house}). These are used for various internal purposes, such as anaphoric resolution, predicate argument typing and grammatical agreement.}

The location of the `standing' event in (\ref{stood-in-house}) is encoded via a \texttt{location\_in} predicate which is formed on the fly from a combination of surface language and automatically identified semantic roles (see \textit{Thematic roles} below). (\ref{stood-in-house-c}) shows this binary predicate with manner and referenced arguments, each encoding their skolem identifiers, space and time. As with all Mephisto structures, this deep predicate can be utilised by Mephisto reasoners for logical inference. The formation of a variety of other spatial predicates follows this principle.

\paragraph{Syntax parser} This module defines and selects hand-crafted grammatically valid syntactic forms to ensure compliance with Lexpresso's controlled English syntax. Selected syntactic structures are described further in Section~\ref{synt-struc}. 

\paragraph{Lexicon} This module is used by the parser to instantiate leaf nodes. The lexicon covers core and domain specific terminology. It includes all major parts of speech containing over 20,000 unique word-forms comprised of a core of circa 12,130 high frequency English tokens plus non-high frequency general and domain specific terms. The class of nouns is comprised of circa 6,900 common nouns, circa 1,000 proper names and 62 forms of pronouns. Nouns are categorised according to certain features: mass/count, number, gender, alienability, and possible syntactic dependencies.

The class of verbs is comprised of main verbs, auxiliaries and modal auxiliaries. There are circa 8,380 main verbs classified according to a number of features, including semantic type, agreement, tense, aspect and mood inflection, syntactic and semantic frames, e.g.~\cite{Fillmore1976,Ruppenhofer2006,Fillmore2012}. 

The class of adjectives is comprised of circa 2,644 forms and categorised into attributive, predicative, comparative and superlative types. Adjectives and other modifiers are further categorised according to the following primarily semantic types: age, amplifier, century, colour, compass, denominal, evaluative, girth, height, noun, objective, ordinal, participle, provenance, religion, shape, size, subjective, weak. This classification is used to stipulate the order adjectives in noun phrases.

Other classes of words consist of articles, cardinal and ordinal numerals, prepositions and other forms such as conjunctions, \textit{wh}-words and directionals.
Finally, there is a special sub-lexicon of domain specific expanded acronyms containing over 41,000 entries. 

The Lexicon and Parser draw on syntactic and semantic knowledge (including \textit{Aktionsart} \cite{Vendler1967}, entity functional capability, and taxonomic relations) to constrain possible interpretations and reduce over-generation of deep CNL structures. 

At this point in the system architecture linguistic content moves from surface to deep Lexpresso modules.

\paragraph{Grapher \& Ambiguity handler} The Grapher transmutes parser outputs into graph structures so associations and semantic structures can be more easily reviewed than with parser output. In cases of multiple interpretations, e.g. (\ref{dec-tr-loc-adjct}), a separate module identifies user defined interpretation preferences. This selects top ranked interpretations in a given context. Where multiple interpretations are equally ranked a separate graph structure is generated for each. The user can compare and select the desired interpretation to ensure each semantic form passed to the reasoner is unambiguous. Evaluation of this potentially burdensome method is required.

\paragraph{Semantic translator \& Thematic roles}
\label{thematic-roles}
The graph structure is then converted into our universal semantic constructs for use by automated inferencing modules, not discussed here. Thematic roles and other linguistic information are identified by combining lexical and constructional semantics from the Syntax parser, Lexicon and Aktionsart modules with the Linguistic Knowledge Base. The results of this process ensure deep structures contain the requisite richness of linguistic semantic information for both inferencing and language generation. The Thematic roles module associates possible semantic roles with generic entities at the highest possible level in the Taxonomy. Subsumed entities will inherit the role associated with their genus. 

\paragraph{Context resolver}
Lexical semantic features associated with nouns (such as gender, animacy, cognitive or other capability) are identified via a relatively shallow hierarchy and used to resolve anaphoric pronouns and \textit{wh}-forms. Types of NP anaphors include personal, reflexive, reciprocal and indefinite pronouns, as well as demonstratives. The only current VP anaphor is forms of the generic verb `do'. A number of rules (not discussed here) determine how anaphors are resolved.

Where possible the CNL Effector (see Figure~\ref{fig:Lexpresso-system-diagram-v3}) makes use of existing Lexpresso modules as already described (such as the lexicon and parser and associated semantic knowledge) to handle the generation of surface language from deep Mephisto semantic constructs. Space constraints prohibit further explanation here.  

\section{Syntactic structures}
\label{synt-struc}
This section exemplifies selected basic syntactic structures permitted by the parser. Space constraints prohibit a comprehensive exposition of all of Lexpresso's syntax. The expository emphasis is on giving a sense of Lexpresso's expressiveness.

\subsubsection{Sentence types}
\label{sentence-types} These include declaratives, interrogatives, directives and indirect speech acts.

\paragraph{Declaratives} These sentence types include basic intransitives (\ref{dec-intr}), transitives (\ref{dec-tr-loc-adjct}) and ditransitives (\ref{dec-ditr}) with and without adjuncts.
\begin{exe}
\ex\label{declarative}				
 \begin{xlist}
	\ex\label{dec-intr} The boy slept on Monday.
	\ex\label{dec-tr-loc-adjct} The woman in the car read the message on the sign.
	\ex\label{dec-ditr} The woman gave the man the document.
\end{xlist}
\end{exe}
\vspace{-0.5cm}

\paragraph{Interrogatives} These can query for a range of syntactic elements: the subject (\ref{inter-subj-ditrans}), object (\ref{inter-obj-trans}) or predicate, such as generic `do', (\ref{inter-pred}) of the basic sentences, as well as temporal information (\ref{inter-temp}) and locational adjuncts (\ref{inter-loc}). Indefinite pronouns can be used to query for any argument; (\ref{inter-yn-indef-pron}) demonstrates its use in a yes/no query.

\begin{exe}
\ex\label{interrogative}				
 \begin{xlist}
	\ex\label{inter-subj-ditrans} Who gave the document to the boy?
	\ex\label{inter-obj-trans} What did the woman read?
	\ex\label{inter-pred} What did the boy do?
	\ex\label{inter-temp} When did she read it?
	\ex\label{inter-loc} What region is she in?
	\ex\label{inter-yn-indef-pron} Did anyone see the woman?
\end{xlist}
\end{exe}
\vspace{-0.5cm}

\paragraph{Directives} These are currently limited to commands to the system to generate situation reports on specified tracks monitored by a track sensor module (not discussed here), e.g. `Show merchant ship situation report on MR41\_PAN-EAV' and `Show commercial aircraft situation report on NAT57\_FL310'. The range of useful commands to the system will to a certain extent dictate development of other directives. Directives are queries expressed in the imperative mood.

\paragraph{Indirect speech} These cover statements with embedded speech act verbs, such as `say' and `tell', e.g. `Michael said that the woman read the document.' and `Michael told Kerry that the woman read the document.' Subclauses introduced by `that' are also permitted in other sentence types, not exemplified here.

\subsubsection{Noun Phrases}
\label{nps}
These can be highly complex with multiple layers of embedding and recursion. The tree structures in (\ref{np}) cover two primary basic types of noun phrase: (\ref{np}a) specifies nouns with pre and post modifiers, whereas (\ref{np}b) specifies conjoined noun phrases. Note its recursion.

\begin{exe}
\qtreecenterfalse
\ex\label{np}
  a. \hskip -0.5cm\small\Tree [.NP [.ENP [.DET the ] [.NP2 [.PRE\\MOD old ] [.N\\\{\textsc{common},\\\textsc{proper}\} man ] [\qroof{from Blueland}.POST\\MOD ] ] ] ]
	\hskip -0.5cm\normalsize
	b. \hskip -0.5cm\small\Tree [.NP [.NPC ENP [.CONJ and ] [.NPC ENP CONJ \qroof{\ldots}.NPC ] ] ] 
\end{exe}

\normalsize
The modifier node is itself internally complex and permits modification by complements and adjuncts, as in (\ref{mods}). Justifications are not given here.

\begin{exe}
\qtreecenterfalse
\ex\label{mods}
 a. \small\Tree [.MOD [.COMP several ] [.MOD friendly ] ]
 \hskip 1.5cm\normalsize
 b. \small\Tree [.MOD [.ADJUNCT {some ancient} ] [.MOD old ] ]
\end{exe}
	
\normalsize
Finally, the trees in (\ref{gen-det}) exemplify permitted noun phrases with genetive-\textit{s}. Again note the recursion in (\ref{gen-det}a).

\begin{exe}
\qtreecenterfalse
\ex\label{gen-det}
 a. \small\Tree [.NP [.GEN-DET [\qroof{the sick woman}.NP ] [.GEN 's ] ] !\qsetw{3cm} [.N house ] ]
 \hskip 1.5cm\normalsize
 b. \small\Tree [.NP [.GEN-DET [.PROP-N Dale ] [.GEN 's ] ] !\qsetw{3cm} [.N car ] ]
\end{exe}
\vspace{-0.6cm}
\normalsize

\section{Semantic structures}
\label{sem-struc}
Kuhn \cite{Kuhn2013} identified some five expressiveness features of Controlled Natural Languages (see a--e in \textit{Fairly high expressiveness} in Section \ref{classif-lexpresso} below). I briefly exemplify these with I and C forms for each semantic type. 

\paragraph{Universal quantification over individuals} Instances of universally quantified entities without an article (\ref{univ-pres-i}) are rendered with an \texttt{all} predicate (\ref{univ-pres-c}) referencing its universally quantified Skolem constant together with the list of relevant linguistic features. Numerals are converted into set operations and can quantify all argument positions and predicates, e.g. `Three men read four documents twice'.

\begin{exe}
\ex\label{univ-pres}				
 \begin{xlist}
\ex[\label{univ-pres-i}I:]{Women stand.}
\ex[\label{univ-pres-c}C:]{\small\texttt{all([skc2],woman(@(skc2,t\_3,s\_2),[female,plural,...])}\\
\small\texttt{=> stands(@(skc2,t\_3,s\_2),[general\_habitual,...])).}}
	\end{xlist}
\end{exe}

\paragraph{Binary or higher relations} In principle Lexpresso does not place a restriction on the arity of relations (\texttt{reads(x,y)} in (\ref{univ-quant-all-arg-c}) exemplifies a binary predicate). However, our ability to reason with higher arity relations is determined by the reasoners used, not discussed here. Our reasoner does not restrict arity either and indeed allows atomic propositions to occur as relation arguments.

\paragraph{General rule structures} Multiple universal quantification can target all argument positions of relations, as in (\ref{univ-quant-all-arg}).

\begin{exe}
\ex\label{univ-quant-all-arg}
	\begin{xlist}
	\ex[\label{all-women-i}I:]{All women always read all documents.}
	\ex[\label{univ-quant-all-arg-c}C:]{\small\texttt{all([skc81,skc82,t\_81],((woman(@(skc81,t\_81,s\_81),[...]) \&}\\
	\small\texttt{document(@(skc82,t\_81,s\_82),[...]))}\\
	\small\texttt{=> reads(@(skc81,t\_81,s\_81),@(skc82,t\_81,s\_82),[...]))).}}
	\end{xlist}
\end{exe}

\textit{If--then} conditionals are also expressible, currently with the form in (\ref{if-then-condit-2}). Note that all argument positions of these conditionals can be universally quantified, as in (\ref{if-then-condit-2}) with `\texttt{all}'.

\begin{exe}
\ex\label{if-then-condit-2}
	\begin{xlist}
	\ex[\label{if-all-women-i}I:]{If all women did not see the car then all women did not see the driver.}
	\ex[\label{x-c}C:]{\texttt{all([skc81],((woman(@(skc81,t\_81,s\_81),[...]) \&}\\
	  \texttt{car(@(skc82,t\_81,s\_82),[...])) =>}\\
		$\mathtt{\sim}$\texttt{sees(@(skc81,t\_81,s\_81),@(skc82,t\_81,s\_82)))) =>}\\
		\texttt{all([skc81], ((woman(@(skc81,t\_81,s\_81),[...]) \&}\\
		\texttt{driver(@(skc84,t\_81,s\_84),[...])) =>}\\
	  $\mathtt{\sim}$\texttt{sees(@(skc81,t\_81,s\_81),@(skc84,t\_81,s\_84),[...]))).}}
	\end{xlist}
\end{exe}

\paragraph{Negation} Weak negation is expressed with a negation operator appended to the front of the negated predicate, as in (\ref{not-read}). This can be applied to any proposition. Strong negation is expressed via lexical negators such as `dislike', `distrust', etc.
\begin{exe}
\ex\label{not-read}
	\begin{xlist}
	\ex[\label{not-read-i}I:]{The woman did not read the document.}
	\ex[\label{not-read-c}C:]{\texttt{woman(@(skc81,t\_22,s\_81),[definite,...]),}\\
	\texttt{document(@(skc07,t\_22,s\_07),[definite,...]),}\\
	$\mathtt{\sim}$\texttt{reads(@(skc81,t\_22,s\_81),@(skc07,t\_22,s\_07),[past,...]).}}
	\end{xlist}
\end{exe}

\paragraph{Second-order universal quantification} This was exemplified in (\ref{univ-quant-all-arg}) in which the predicate `read' is universally quantified by `always' and rendered as an operator over the time of the predicate.

Other features articulated by Kuhn as determinants of expressiveness were existential quantification, as in (\ref{stood-in-house}), equality, as in (\ref{equality}), and types of speech acts (not exemplified due to space constraints but mentioned in Section~\ref{synt-struc}). 
 
\begin{exe}
\ex\label{equality}
	\begin{xlist}
	\ex[\label{identical-i}I:]{Andrew White is the Prime Minister.}
	\ex[\label{identical-c}C:]{\small\texttt{Andrew\_White(@(skc6,t\_10,s\_6),[...]),}\\
	\texttt{prime\_minister(@(skc7,t\_10,s\_7),[...]),}\\
	\texttt{identical[@(skc6,t\_10,s\_6),@(skc7,t\_10,s\_7)].}}
	\end{xlist}
\end{exe}

\paragraph{Discourse structures}
\label{discourse}
The paragraph is taken as the basic unit of discourse. For the purpose of anaphoric resolution, a new paragraph signifies a new discourse context. A single sentence can constitute a paragraph. Anaphora occurs within a discourse unit.

\section{Classification of Lexpresso}
\label{classif-lexpresso}
Kuhn \cite{Kuhn2013} presented a classification scheme for CNLs labelled with the acronym PENS. This classifies a CNL according to a five-tier ranking (with 5 for maximal) for each of four orthogonal categories of Precision, Expressiveness, Naturalness and Simplicity. Based on my assessment of Lexpresso against Kuhn's `PENS' scheme, I tentatively classify it as P^{3-4} E^{4} N^{4-5} S^{3} as evidenced by the following paragraphs. 

\paragraph{Precision---reliably \& semi-deterministically interpretable} P^{3-4} Although it is not currently possible for any natural English language text to be deterministically transformed by Lexpresso into a formal logic representation, the syntax is heavily restricted enough to make automatic interpretation reliable. In cases where the natural language syntax is ambiguous and not automatically disambiguated, multiple interpretations are presented. The user is then consulted to select the desired interpretation. Once done, controlled natural language is deterministically translated into formal structures. There is also a well established, conceptually broad, underlying formalism.

\paragraph[\label{expressiveness}]{Fairly high expressiveness} E^{4} The range of propositions that Lexpresso can express includes all those articulated by Kuhn: (a) universal quantification over individuals; (b) relations of arity greater than 1; (c) general rule structures (if-then conditionals with multiple universal quantification that can target all argument positions of relations; (d) negation (strong negation or negation as failure); (e) general second-order universal quantification over concepts and relations; (f) existential quantification; (g) equality; and (h) types of speech acts including declarative, interrogative, directive and indirect. See Section~\ref{sentence-types} for examples.

\paragraph{Fair degree of naturalness} N^{4-5} While large scale texts have not been written in the language, small fairly natural texts and spoken interactions with internal interdependencies are parsable.

\paragraph{Simplicity} S^{3} Lexpresso can be exactly, comprehensively defined with accepted grammatical and logical notations but it is likely to require more than ten pages to describe all its syntactic and semantic properties.\footnote{The description of Lexpresso's features presented here is not considered comprehensive and therefore does not qualify as an indicator of its simplicity score.}

\section{Conclusion}
\label{concl}
This brief introduction to the Controlled Natural Language---Lexpresso---has presented the system architecture, and exemplified its main syntactic and semantic features. These features have been compared to Kuhn's \cite{Kuhn2013} PENS classification system. Against this comparison I have tentatively classified Lexpresso as a P^{3-4} E^{4} N^{4-5} S^{3} CNL. 
According to this classification Lexpresso is a reliably or perhaps deterministically interpretable language, with high expressiveness, considerable naturalness and would require a lengthy treatment to cover its syntax and semantics. Given the page limit, this paper has not in any detail discussed Lexpresso's limitations nor the tight integration with the knowledge representation and reasoning capabilities which constitutes a significant component of our high-level information fusion system for which Lexpresso functions as a natural language interface. 

\subsubsection{Acknowledgements.} I thank Dale Lambert, Kerry Trentelman, Andrew Zschorn and Takeshi Matsumoto for discussions on issues raised in this paper and the first two named plus Nathalie Colineau and three anonymous reviewers for valuable comments on an earlier version. Both Andrew Zschorn and Takeshi Matsumoto have contributed to the development of Lexpresso. The final publication is available at link.springer.com

\bibliographystyle{splncs} 
\bibliography{Saulwick-Lexpresso-2014-06-30}

\end{document}